\documentclass[a4paper,10pt,twoside]{llncs}

\pdfoutput=1

\usepackage[english]{babel}
\usepackage[latin1]{inputenc}       
\usepackage{graphicx}
\usepackage{moreverb}               
\usepackage{alltt}                  
\usepackage{amssymb}
\usepackage{algorithmic}
\usepackage[chapter,boxed]{algorithm}
\usepackage[square,numbers]{natbib} 

\usepackage[breaklinks,colorlinks,linkcolor=black,citecolor=black,pagecolor=black,urlcolor=black,anchorcolor=black]{hyperref}

\hypersetup{pdfauthor=Stasinos Them. Konstantopoulos}
\hypersetup{pdftitle=A Data-Parallel Version of Aleph}
\hypersetup{pdfkeywords=ILP inductive logic programming aleph data-parallel MPI}

\makeindex

\newcommand{\todo}[1]
{
\begin{quote}
\textbf{COMMENT:}
     #1
\end{quote}
}
\renewcommand{\todo}[1]{}

\newcommand{\disappear}[1]{}

\newcommand{\term}[1]{\emph{#1}}

\newcommand{\prog}[1]{\mbox{\path{#1}}}

\newcommand{\quoted}[1]{`#1'}
\newcommand{\quotes}[1]{`#1'}

\newenvironment{centre}  {\begin{center}}  {\end{center}}
\newenvironment{progenv} {\begin{alltt}}   {\end{alltt}}

\begin{document}

\title{A Data-Parallel Version of Aleph}
\author{Stasinos Th. Konstantopoulos}
\institute{Alfa-Informatica, Rijksuniversiteit Groningen
  \\ Postbus 716, 9700AS Groningen, The Netherlands
  \\ Telephone: (+) 31-50-3635935
  \\ Fax: (+) 31-50-3636855
  \\ \email{konstant@let.rug.nl}
}

\maketitle

\begin{abstract}
  This is to present work on modifying the Aleph ILP system so that it
  evaluates the hypothesised clauses in parallel by distributing the 
  data-set among the nodes of a parallel or distributed machine.
  
  The paper briefly discusses MPI, the interface used to access
  message-passing libraries for parallel computers and clusters.  It
  then proceeds to describe an extension of YAP Prolog with an MPI
  interface and an implementation of data-parallel clause evaluation
  for Aleph through this interface. The paper concludes by testing the
  data-parallel Aleph on artificially constructed data-sets.
\end{abstract}

Inductive Logic Programming (ILP) is the Machine Learing discipline
that deals with the induction of Logic Programmes from examples.
Despite the constant advances in computing hardware, ILP remains a
computationally expensive application with learning sessions taking
many hours or even days to complete, even on the most powerful
processors. A four-day experiment that had to be repeated all over
again because of some minor bug in the background knowledge or a
powerful server brought to the thrashing point by a memory-hungry
ILP experiment has to be accepted as a fact of life.
This makes ILP algorithms good candidates for parallel or distributed
computing, so that multiple CPUs or workstations can share the
computational and data-storage load of an ILP experiment.

The Message Passing Interface (MPI) is a library specification for
message passing between the nodes of a parallel machine or workstation
cluster. MPI libraries, thus, facilitate communication between
the processes involved in a parallel computation.

This paper describes an extension of the Yap Prolog compiler
with an interface to MPI libraries (Sections~\ref{paral:mpi}
and~\ref{paral:yap}) and an adaptation of the Aleph ILP
system so that it can evaluate hypothesised clauses in parallel
through this interface (Section~\ref{paral:aleph}).
Section~\ref{paral:conclusions} concludes by addressing
the issue of what kinds of problems this particular kind of
parallelism is suitable for.

\section{The Message Passing Interface}
\label{paral:mpi}

The \term{Message Passing Interface} (MPI) is a specification for the
Application Programmer's Interface (API) to libraries that
facilitate datagramme-style communication between the
processors of a parallel machine or workstation cluster.
What is meant by datagramme-style communication is that the
information is transmitted in packets rather than through pipes,
although the actual transmission is typically synchronous.

MPI should not be confused with libraries implementing parallel
numerical methods, or with parallelising
compilers. MPI provides the message-passing infrastructure
necessary for the communication between the nodes of a parallel
computation, and does not automate in any way the actual
parallelisation of the code as, for example, a parallelising
compiler would.

It should also be pointed out that MPI is not a library, but an API
specification. The advantage of conforming to the MPI specification is
that programmes can link to any MPI library without modifications,
allowing for greater portability between all kinds of varied and
diverse parallel architectures.
In the remainder of this section, the MPI functions that are
pertinent to the implementation of the MPI version of Aleph
will be introduced. For a more complete description of MPI, the reader
is referred to the MPI standard defined and maintained by the MPI
Forum \citep{mpi:mpi, mpi:mpi2} or user's guides to either MPI in
general or some particular MPI library
\citep{gropp:1996,pacheco:1998}.

\subsection{Basic MPI Concepts}

It was mentioned above that MPI facilitates the communication between
the nodes of a parallel architecture, but it would be more accurate to
say that it facilitates the communication between the processes
involved in a parallel computation. In order to start a program that
is using an MPI library on a parallel machine, a separate,
architecture-specific mechanism --- a job-queueing system, for example
--- has to load identical copies of the program onto each and every
node. The MPI library will then provide the methods for passing
messages between these processes abstracting away from the
architecture of the machine, so that they can be running on
the nodes of a parallel computer, the workstations of a network, or
even be multiple processes running on the same processor. It is of
obvious benefit to spread the processes as evenly among the available
resources as possible, but that is not part of the MPI protocol: it is
the queueing system's task to start the processes and enforce its
queueing policy. For the remainder of this paper, \term{process}
and \term{node} will be used interchangeably to refer to a node of
the computation at the abstract, MPI level, regardless of how that
maps to the actual processor nodes of the hardware.

The processes involved in a parallel computation are identified by
\term{communicator} and \term{rank}. A \term{communicator} is a
collection of nodes involved in a sub-task of the computation, and
a computation might keep all the available nodes within one
communicator or split them among several. The goal of MPI is to
facilitate fast and efficient \emph{intra}-communicator communication,
whereas \emph{inter}-communicator communication is meant to be sparser
and not as performance-critical.

Within each communicator, processes are identified by their
\term{rank}, which is an enumeration of the processes, starting from
0. The node with rank~0 will be called the \term{head node}. MPI
defines \prog{MPI_COMM_WORLD} to be the global communicator that
groups together all the nodes, useful for applications that do not
need to group their nodes into separate communicators. Since this is
case here, for the remainder of this paper all references to rank
will refer to the node by that rank in the global communicator, and
the communicator argument to MPI functions will not be shown, since it
would invariably be \prog{MPI_COMM_WORLD}.

The most basic operation that MPI facilitates is the point-to-point
sending and receiving of a message. A message consists of an array of
data, a \term{type}, and a \term{tag}. The \term{type} is one of
several predefined data-types supported by MPI. All the the usual C
data-types, like characters, integers, and floating-point numericals,
are supported. It should, however, be noted that although MPI types
are stored locally according to the native binary representation for
that type, all the appropriate conversions\footnote{For example
  between ASCII and EBCDIC characters, or 1's-complement and
  2's-complement integers.} 
are carried out when typed data is been transmitted, thus making it
possible to spread a computation over heterogeneous workstation
clusters.

Lastly, a message carries a numerical \term{tag} which can be
interpreted as
a message type. Messages with different tags \quotes{live} in a
different space, and a receive action must specify (by tag) which
sorts of
messages it should be allowed to receive; the messages carrying any
other tag are to be ignored. This mechanism allows for a certain
degree of asynchroneity, as communications of different
\quoted{sorts} can be kept apart without having to rely on
synchronising the sender and the receiver.

With the above concepts in mind, the elementary MPI operators will
have the following C declarations:
\begin{progenv}
int MPI_Send(message, count, datatype, dest, tag)
void *message;
int count, dest, tag;
MPI_Datatype datatype;

int MPI_Recv(message, count, datatype, source, tag, status)
void *message;
int count, source, tag;
MPI_Datatype datatype;
MPI_Status *status;
\end{progenv}
where \prog{MPI_Send()} would dispatch \prog{count} bytes from memory
location \prog{message} to the node of rank \prog{dest}. To receive
the message, the recipient must issue an \prog{MPI_Recv()} specifying:
the maximum number of bytes to accept and where to place them;
the source node's rank or \prog{MPI_ANY_SOURCE};
the message's type and
tag (or \prog{MPI_ANY_TYPE} and \prog{MPI_ANY_TAG}, respectively);
and the memory location where the status of the transfer should be
stored. This last \prog{MPI_Status} structure
includes information such as the actual message length, type and tag.

\begin{figure}

\begin{centre}
\begin{boxedverbatim}
if(my_rank == 0) {
  MPI_Send("Hello World", 12, MPI_CHAR, 1, 0);
  MPI_Send("Hello World", 12, MPI_CHAR, 2, 0);
}
else {
  char buf[255];
  MPI_Recv(buf, 255, MPI_CHAR, 0, 0);
  puts(buf);
}
\end{boxedverbatim}
\end{centre}

\caption{Example usage of \texttt{MPI\_Send()} and \texttt{MPI\_Recv()}}
\label{fig:paral:send_recv}
\end{figure}

The C code fragment of Figure~\ref{fig:paral:send_recv} demonstrates
this simplest form of message passing, where the processes with rank~1
and~2
receive a string from the head node, and then print it out.
The message is marked as being of type \prog{MPI_CHAR} (character
array) and tagged with \prog{0} by the sender, and the receivers are
also specifying that they are only accepting messages of that type and
tag. The
third, fourth and fifth argument of \prog{MPI_Recv()} could have also
been \prog{MPI_ANY_TYPE}, \prog{MPI_ANY_SOURCE} and
\prog{MPI_ANY_TAG}, to mean that they would accept
messages of any type, any tag, and from any sender respectively.

The \prog{MPI_Send()}/\prog{MPI_Recv()} functions described above are
synchronous, in the sense that once \prog{MPI_Send()} is called, the
caller is blocked until a matching \prog{MPI_Recv()} is issued and
completed (and, of course, vice versa). The MPI protocol also allows
for asynchronous communication through the \term{immediate} series of
point-to-point communications, which return immediately instead of
blocking until the transfer is completed and can thus take advantage
of hardware that facilitates memory transfers independently of the
computations performed in the CPU. Since these functions have not been
employed in Aleph/MPI, they will not be dealt with here any further.

\subsection{Some More MPI Functions}

The strength of the MPI specification is that it provides
higher-level operators which, although possible to implement with the
more basic functions described above, can be implemented more
efficiently by architecture-specific code. This way parallel
programmes employing MPI can at the same time be highly portable
between architectures and optimised for the architecture
they are running on, assuming the existence of a native MPI
implementation.

One such function is the \term{broadcast} function, which allows one
node (the \term{root} of the broadcast) to send a message to all other
processes. Broadcasting is used when a new or updated data structure
needs to be propagated through all the nodes. Broadcasting is
synchronous, which is to say that \emph{all} the nodes have to reach
the point in their code where the broadcast call is made before a
broadcast can be successfully completed.

Broadcasting is performed with the \prog{MPI_Bcast()} function:
\begin{progenv}
int MPI_Bcast(message, count, datatype, root);
void *message;
int count, root;
MPI_Datatype datatype;
\end{progenv}
which must be invoked by \emph{all} processes.
It should be stressed at this point that the \prog{MPI_Bcast()} calls
will block the calling process until all processes have issued a
\prog{MPI_Bcast()} and the broadcast (or at least a process's
involvement in the broadcast) has been completed.
As is typically the case with synchronous communication, it is the
application programmer's responsibility to ensure that all the nodes
will reach the \prog{MPI_Bcast()} call at approximately the same time,
so that as few CPU cycles as possible are waisted while the nodes that
issued the \prog{MPI_Bcast()} call first are waiting for the rest.

Furthermore, all the nodes of the broadcast need to know in advance
which is the root of the broadcast and provide its rank as the
\prog{root} argument. It should also be noted that there is no
notion of message tags, so that a matching root is the only
requirement for delivering a broadcast message.

Using \prog{MPI_Bcast()}, the code snippet of
Figure~\ref{fig:paral:send_recv} would be re-written as in
Figure~\ref{fig:paral:bcast} (enriched with some initialisation,
finalisation and environment-retrieving functions explained below). It
can be seen there that all nodes make the \prog{MPI_Bcast()} call and
they all know in advance which is the root of the broadcast, the type
of the message, and what size buffer will be sufficient for receiving
it. The difference between the root node and the receipient nodes is
that the root node populates the buffer \prog{buf} with some data and
then issues the broadcast, whereas the receipients wait for the
broadcast to fill in the buffer, so that they can subsequently make
use of the information found there.

The broadcast call is one of the functions that demostrate the power
of MPI: the API of the \prog{MPI_Bcast()} call is the same no matter
whether it is portably implemented as a series of send and receive
operations, or some more efficient architecture-specific way; for
example the message could be placed in shared memory and copied from
there by all the processes in architectures that support shared
memory.

\begin{figure}

\begin{centre}
\begin{boxedverbatim}
char buf[255];

MPI_Init(&mpi_argc, &mpi_argv);
MPI_Comm_rank(&my_rank);

if(my_rank == 0) { strcpy(buf, "Hello World"); }

MPI_Bcast(buf, 12, MPI_CHAR, 0);

if(my_rank != 0) { puts(buf); }

MPI_Finalize();
\end{boxedverbatim}
\end{centre}

\caption{Example usage of \texttt{MPI\_Bcast()}}
\label{fig:paral:bcast}
\end{figure}

Some more functions that need to be mentioned are the initialisation
and finalisation functions, as well as functions that provide
information about the parallel environment. \prog{MPI_Init()} accepts
as argument the process's argument vector and initialises the system
with the number of processes specified there. After initialisation,
\prog{MPI_Comm_size()}, for example, can be used to retrieve the
number of processes in the communicator, \prog{MPI_Comm_rank()} to
retrieve the calling node's rank, and so on.  \prog{MPI_Finalize()} is
called without any arguments to terminate the computation.

Finally, one should note \prog{MPI_Probe()} and \prog{MPI_Get_count}:
\begin{progenv}
int MPI_Probe(source, tag, status);
int source, tag;
MPI_Status *status;

int MPI_Get_count(status, datatype, count);
MPI_Status *status;
MPI_Datatype datatype;
int *count;
\end{progenv}
\prog{MPI_Probe()} can be used to receive data of unknown size, by
\quotes{probing} the queue for information without actually extracting
any data. \prog{MPI_Probe()} populates the \prog{status} structure
with information about the head message in the queue with matching
\prog{source} and \prog{tag}. \prog{MPI_Probe()} will block (just like
\prog{MPI_Recv()} if there is no matching message yet.

The size of the message received, however, is not directly available
as a field of the \prog{status} structure and a call to
\prog{MPI_Get_count()} is required to extract it.
\prog{MPI_Get_count()} also expects the datatype as which the message
will be interpreted to be provided.

\section{The Yap/MPI Interface}
\label{paral:yap}

Yap is a Prolog system developed at the University of Porto and at the
Federal University of Rio de Janeiro \citep{costa:yap}.
Yap includes two
mechanisms for extending the library with foreign predicates,
typically written in C: one static (through libraries or object-code
files linked into the bulk of Yap's code) and one dynamic (through
dynamic libraries, linked at run-time). The Prolog interface to MPI
described here consists of a static extension to the Yap library. This
approach was chosen because MPI installations might sometimes only be
available as static libraries\footnote{As is the case with both the
  MPI implementations installed on the Linux-workstation cluster of
  the University of Groningen where Yap/MPI was tested.},
forcing the interface code to be static as well.

It should be stressed that what is being described in this paper is
not parallelising or in any way
modifying any of the logical aspects of Yap, and, indeed, no changes
have been made to either the abstract machine implementation or the
internal database mechanism. Just like MPI itself is not a
parallelising compiler but only a message-passing mechanism, a Prolog
interface to MPI only provides the infrastructure for passing
messages between the nodes of a parallel computation. The interface is
implemented as an additional foreign library and the only changes made
within the existing Yap code were are at the initialisation routine,
where the \prog{mpi_*} predicates are declared and the 
MPI-related command-line arguments extracted and stored so
that they can be used by \prog{mpi_open/3}.

\subsection{Prolog Term Messages}

Prolog operates at a higher level with respect to data structures than
C and MPI, which means that a Prolog interface to MPI cannot let the
arguments of Prolog predicates simply fall through to the C calls it
is based on, but it has to re-express them as one of the MPI
elementary data-types

In Prolog all data is expressable as terms, which terms are (at the
level of the Prolog interface) treated uniformly regardless of whether
they contain integral, real, alphanumerical, or any any other kind of
data as arguments. This means that the MPI interface should be able to
transmit unresticted Prolog terms, the binary representation of which
might be radically different between different machines, making it
impossible to simply clone the binary term and graft it in the
receiving machine's memory.

For this reason it is the string representation of terms that is
transmitted, which is then parsed back into a term on the receiving
node. This was achieved by reusing the code of the term parser and
printer available in the I/O module of the Yap system; on the sending
side terms are translated into their string representation and then
transmitted as arrays of type \prog{MPI_CHAR}. On the receiving side,
they are parsed back into the internal representation and the
resulting Prolog term used to bind an \quotes{output} variable.

\subsection{Point-to-Point Communication}

The \prog{mpi_send/3} and \prog{mpi_receive/3} predicates are
implementing the interface to MPI's synchronous, point-to-point
\prog{MPI_Send()} and \prog{MPI_Recv()}. They have the
following semantics:
\begin{progenv}
   mpi_send(+Data, +Destination, +Tag)
mpi_receive(-Data, ?Source,      ?Tag)
\end{progenv}
where \prog{mpi_receive/3} binds the \prog{Data} variable to the term
sent by \prog{mpi_send/3} if the \prog{Source} and \prog{Tag}
variables can be unified with those of the corresponding
\prog{mpi_send/3} activated.
Figure~\ref{fig:paral:send_recv_pl} demonstrates the usage of
\prog{mpi_send/3} and \prog{mpi_receive/3}; it is equivalent to the
\quotes{Hello World} fragment in Figure~\ref{fig:paral:send_recv}.

\begin{figure}

\begin{centre}
\begin{boxedverbatim}
paral(0):- !,
     mpi_send('Hello World', 1, 0),
     mpi_send('Hello World', 2, 0).
paral(_):-
     mpi_receive(Message, 0, 0),
     writeq(Message).

:- mpi_open(Rank, NumProc, NameProc),
     paral(Rank),
     mpi_close.
\end{boxedverbatim}
\end{centre}

\caption{Example usage of \texttt{mpi\_send/3} and \texttt{mpi\_receive/3}}
\label{fig:paral:send_recv_pl}
\end{figure}

The \prog{Data} argument in
\prog{mpi_receive/3} must be an unbound variable rather than a
partially or even fully instantiated term.
It would have been possible to allow this argument to be partially or
fully instantiated, and then simply have the predicate fail if the
argument fails to unify against the term that has been received, but
that would have been misleading: once the source and tag arguments
match, the message will be extracted from the message queue and only
then unified with \prog{Data}. Since there is no way to push messages
back into the head of the queue, the only reasonable design choice is
to always accept a message if the tag and source match, in other
words require that the first argument of \prog{mpi_receive/3} is an
unbound variable.

To make this point clearer, consider the two variations of the code of
Figure~\ref{fig:paral:send_recv_pl} shown in
Figure~\ref{fig:paral:send_recv_pl2},
where the message is encapsulated
in a \prog{msg/2} term which carries a filename to output to as well
as the message itself. The receiving nodes perform a simple
transformation on the filename sent by the head node and then print
the text to a file by that name.
The (correct) code to the left accepts any term (assuming the
sender and tag match) and then performs the necessary checks,
whereas the code to the right incorrectly assumes that because the
sent message cannot be unified with the \prog{msg(file1,Text)} term it
expects, it will not be extracted from the queue and a second
attempt to receive it can be made. In order to avoid this kind of
confusion, \prog{mpi_receive/3} is implemented so as to immediately fail
without calling \prog{MPI_Recv()} if its first argument is not an
unbound variable.

\begin{figure}

\begin{centre}
\begin{boxedverbatim}
paral(0):- !,
  mpi_send(msg(file1, 'Data'),
           1, 0),
  mpi_send(msg(file2, 'Data'),
           2, 0).
paral(_):-
  mpi_receive(Message, 0, 0),
  (Message = msg(file1, Text),
   File = 'file1.data'
  ;
   Message = msg(file2, Text),
   File = 'file2.data')

  open(File, write, S),
  writeq(S, Text), close(S).

:- mpi_open(Rank, _, _),
  paral(Rank),
  mpi_close.
\end{boxedverbatim}
~
\begin{boxedverbatim}
paral(0):- !,
  mpi_send(msg(file1, 'Data'),
           1, 0),
  mpi_send(msg(file2, 'Data'),
           2, 0).
paral(_):-
  (mpi_receive(msg(file1,
                Text), 0, 0),
   File = 'file1.data'
  ;
   mpi_receive(msg(file2,
                Text), 0, 0),
   File = 'file2.data')
  open(File, write, S),
  writeq(S, Text), close(S).

:- mpi_open(Rank, _, _),
  paral(Rank),
  mpi_close.
\end{boxedverbatim}
\end{centre}

\caption{Example usage of \texttt{mpi\_recv/3} with uninstantiated
  (left) and partially instantiated (right) \texttt{Data} argument.}
\label{fig:paral:send_recv_pl2}
\end{figure}

One thing that also needs to be noted is that the functions
\prog{MPI_Probe()} and \prog{MPI_Get_count()} are used to
retrieve the size of the message before its actual reception. In other
words, a successful \prog{mpi_receive/3} activation translates into
two C-level MPI calls: one to \prog{MPI_Probe()} and one to
\prog{MPI_Recv()}\footnote{not counting the \prog{MPI_Get_count()}
  call, which is definitelly not going to require any inter-process
  communication.}.
This approach was taken in order
to ensure that no prior limit is set on the size of the terms that
will be transmitted. This is particularly important for the
application of the interface described here, since Aleph needs to
transmit potentially enormous lists of examples covered between the
worker nodes and the master node (see section~\ref{paral:aleph}
below). The cost of the \prog{MPI_Probe()} call depends on the
specifics of the MPI implementation used, and can be from negligible
(if message-queue statistics are available locally on the node
issueing the \prog{MPI_Probe()} call) to comparable with a full
\prog{MPI_Recv()}.

An alternative would be to implement \prog{mpi_send/3} and
\prog{mpi_receive/3} in a way that allows for the transmission of
arbitrarily long terms while at the same time transmitting smaller
terms with only one invocation of \prog{MPI_Send()}/ \prog{MPI_Recv()}.
One way of achieving this would be transmitting longer messages as
chains of message packets. This requires sending some extra
\quoted{control information} along with the message (most notably
whether this is the last packet or not), which could be easily
accomplished by, for example, encapsulating the whole message in a
term.
This was not done for this prototype implementation, but might be
worth doing if the MPI interface is to be used for
a more varied range of applications rather than only Aleph, as is
currently the case.

\subsection{Broadcasting}

A similar approach was used for the Prolog interface to
\prog{MPI_Bcast()}, except that in this case there is no tag argument,
since \prog{MPI_Bcast()} does not support message tags:
\begin{progenv}
mpi_bcast(+Data, +Root)
mpi_bcast(-Data, +Root)
\end{progenv}
The first calling mode is for the root node and the second for all the
receiving processes. This is, in fact, enforced by the implementation of
\prog{mpi_bcast/2} by comparing the value of \prog{Root} with the
rank of the node executing the call.

Analogously to \prog{mpi_send/3} and \prog{mpi_recv/3},
\prog{mpi_bcast/2} is also not depending on the user to provide a
maximum message size. This is achieved by implementing
\prog{mpi_bcast/2} as two \prog{MPI_Bcast()} calls, the first of which
is used to transmit the length of the string representation of the
actual term to be transmitted.

Finally, \prog{mpi_open/3} and \prog{mpi_close/0} are used to start
and terminate the parallel computation. \prog{mpi_open/3} uses the
user arguments in the Yap command-line (i.e. the arguments after the
\prog{--} on the Yap command line) as MPI arguments to pass to
\prog{MPI_Init()}, and then it
uses the appropriate functions to retrieve the number of processes and the
current node's rank and name, which it unifies with its three
arguments. \prog{mpi_close/0} accepts no arguments, and simply calls
\prog{MPI_Finalize()}.

To demonstrate the usage of the Prolog interface to
\prog{MPI_Bcast()}, the code fragment in Figure~\ref{fig:paral:bcast}
is given in its Prolog
equivalent in Figure~\ref{fig:paral:bcast_pl}.

\begin{figure}

\begin{centre}
\begin{boxedverbatim}
paral(0):- !,
     mpi_bcast('Hello World', 0).
paral(_):-
     mpi_bcast(Message, 0),
     format(Message, []).

:- mpi_open(Rank, NumProc, NameProc),
     paral(Rank),
     mpi_close.
\end{boxedverbatim}
\end{centre}

\caption{Example usage of \texttt{mpi\_bcast/3}}
\label{fig:paral:bcast_pl}
\end{figure}

\section{Evaluating Clauses in Parallel}
\label{paral:aleph}

Aleph \cite{srinivasan:aleph} is an ILP system written in Prolog. It
implements (among others) the Progol algorithm
\cite{muggleton:1995,muggleton:1994}, a sequential-cover ILP
algorithm.

The Prolog interface to MPI libraries described above, is used to
extend Aleph~3 so that it evaluates in parallel the hypothesised clauses
it builds during the search for a good clause. The predicates within
Aleph that were mostly influenced were those pertaining to loading the
example files (since the examples had to be distributed among the
processes) and the those implementing the example-proving mechanism
itself.

\subsection{Loading the Examples}
\label{paral:load_ex}

In Aleph, the example files are read and the examples asserted in the
internal database (IDB)
as \prog{example/3} terms, with arguments a unique number for each
example, whether it's a positive or a negative example, and the
example itself.
For Aleph/MPI, the examples have to be distributed among the
nodes. This is done by having the head node read the example files in,
assign each example its numerical ID and construct the
\prog{example/3} terms, and transmit to the appropriate node.

One of the advantages of the parallel clause evaluation is that
each node needs keep in memory only the examples that it will be
evaluating each clause, allowing for data sets that wouldn't fit in
any single computer to be employed.

The disadvantage of not keeping all examples on all nodes is that the
work-load might become unbalanced over the course of a learning
session. This is due to the fact that covered positives get removed
from the examples pool, so that only for the very first clause
added to the hypothesis is it guaranteed that the work-load will be
evenly balanced among the nodes. This is, in practice, not as big a
problem as it might seem, since for large numbers of examples it is
expected that the
distribution along the nodes will remain practically even. When the
example pool gets small enough that some nodes might be left with
drastically less work to do, then the bulk of the computation will
have been already performed and the losses will be bounded within a
small fraction of the total time spent. Furthermore this only applies
to the positive examples, since negative examples do not get removed
from the pool and will remain balanced for the duration of the
learning session.

One way to further alleviate this problem is to scramble the example
files, in order to prevent any patterns emerging from the
example-generation process from resulting in the removal of large
numbers of \emph{consecutive} examples by one clause. The removal of
large numbers of examples is, of course, still the goal of the whole
process, but we would rather have the covered examples spread as
uniformly as possible among the nodes.
This is currently done by distributing the examples by modulo: that
is, if there are three workers, the examples
would be distributed like so:
\begin{center}\begin{tabular}{lll}
worker 0 & worker 1 & worker 2 \\
ex~0     & ex~1     & ex~2     \\
ex~3     & ex~4     & ex~5     \\
ex~6     & ...      &          \\
\end{tabular}\end{center}
and so on, so that removing ranges of consecutive example would
equally lighten the workload of all nodes.

\subsection{Proving the Examples}

Aleph/MPI is designed so that one controlling (\term{master}) process
is performing all the non-parallel calculations, distributing work to
the rest of the nodes (the \term{workers}), and collecting and
collating the results.

Besides the addition of the MPI initialisation and
finalisation code and the distributing of examples among the nodes,
the most important differences between Aleph and Aleph/MPI lie under
the \prog{induce/0} and \prog{prove_cache/8} predicates.

\begin{figure}

\begin{center}
\includegraphics{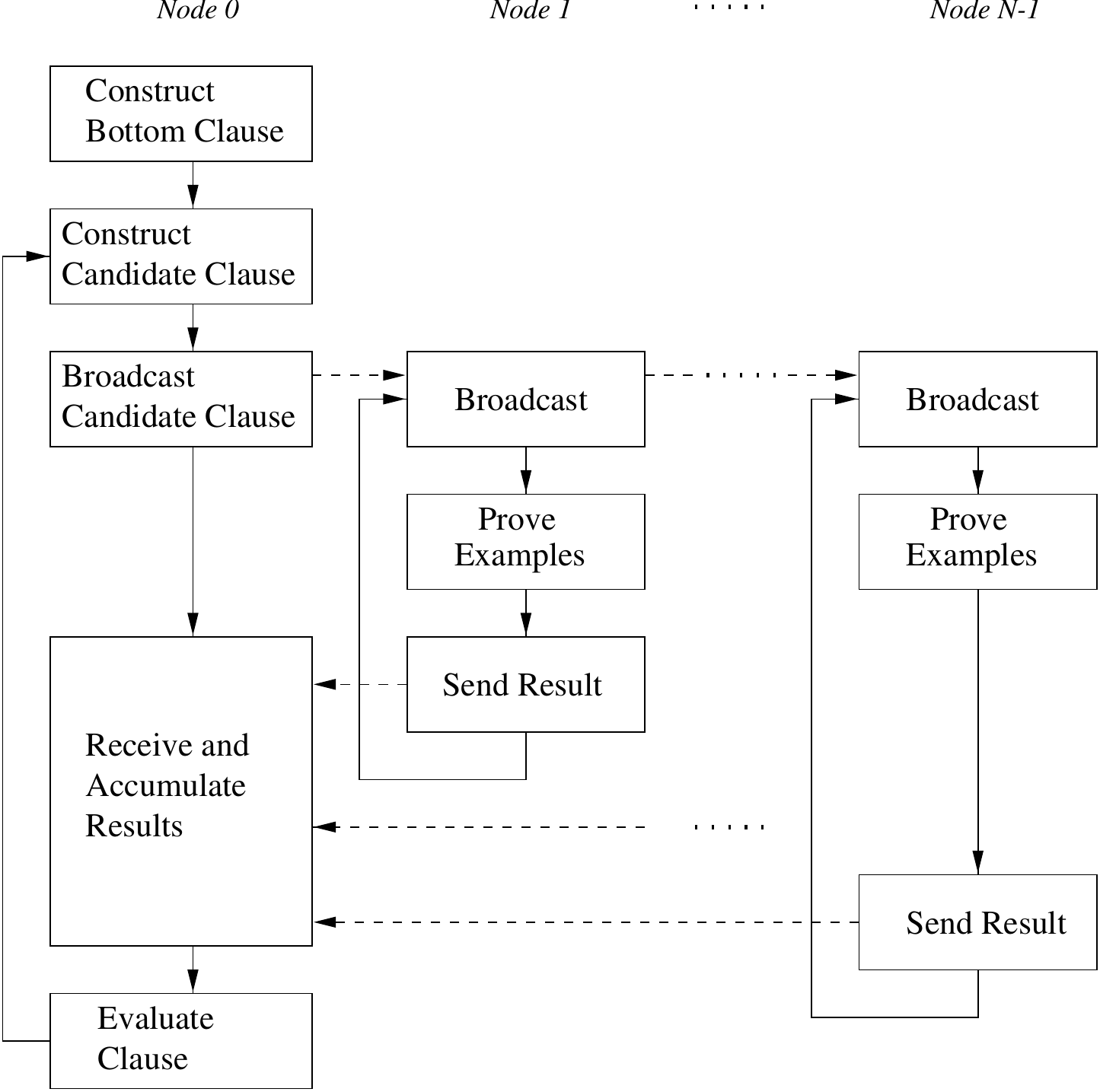}
\end{center}

\caption{A rough schematic of the clause construction and evaluation
  loop: solid lines represent execution flow and dashed lines data
  flow between the nodes.}
\label{fig:paral:loop}
\end{figure}

The \prog{induce/0} predicate is the main
clause construction and evaluation loop, supplemented in Aleph/MPI
by \prog{induce/1 (+Rank)}. \prog{induce/0} retrieves the
rank of the current node and calls \prog{induce/1}, which 
(as can also be seen in Figure~\ref{fig:paral:loop})
behaves
differently for the master node than for the workers:
\begin{enumerate}
\item The head node (master) activates \prog{induce/1} as
  \prog{induce(0)} and has it behave, effectively, the same as the
  original \prog{induce/0}, going through the example saturation,
  clause construction and clause evaluation loop. The difference is
  that the master is now broadcasting to the workers a request for the
  evaluation of a clause, rather than proving the clause itself.  When
  the master receives the answers from the workers, it calculates the
  union of the successful example intervals and the summation of the
  numbers of positives and negatives covered and proceeds (as per
  single-processor Aleph) to apply the evaluation function based on
  these numbers, append successful clauses to the current theory and
  remove covered positive examples from the pool.
\item When activated with any non-zero rank value, \prog{induce/1}
  goes into the workers' loop that issues a broadcast, acts upon prove
  requests as soon as they get broadcast, uses \prog{mpi_send/3} to
  transmit back to the master the list of successful examples, and
  returns to waiting for the next broadcast.
\end{enumerate}
It should be noted that it is the master's responsibility to keep
track of the \quotes{active} (not covered yet) positive examples,
since it is in the master node that the decision whether to append a
clause to the theory or not is made. This means that the master has to
keep a list of the active example indexes (but not the examples
themselves) and include this list in the prove request along with the
clause that needs to be evaluated. A performance improvement could be
gained by keeping this list local to the workers and issuing the
appropriate update request every time a clause gets accepted into the
theory, under the assumption that:
\begin{enumerate}
\item There will be a lot more unsuccessful clauses than successful
  ones.
\item There is a significantly higher cost in transmitting the
  examples-to-prove list, versus the cost of the update requests.
\end{enumerate}
The second assumption might not be always satisfied, since it is the
case that in modern workstation clusters it is the delay of
establishing a connection between nodes that is responsible for the
transmission costs, rather than the low bandwidth of the network.

The \prog{prove_cache/8} predicate is the entry point to the
example-proving mechanism: it first checks to see if a given clause
has already been proven (and cached), and if yes returns the already
calculated and cached coverage, otherwise it tries to prove the
examples with this clause and returns (and caches) the results.

Aleph/MPI parallelises the example proving mechanism,
so the differences between Aleph and Aleph/MPI can be hidden beneath
\prog{prove_cache/8}.
In other
words, the changes made to the predicates employed by 
\prog{prove_cache/8} as well as within \prog{prove_cache/8} itself,
are not visible when activating \prog{prove_cache/8}, which retains
the same semantics as with Aleph:
\begin{progenv}
prove_cache(+Mode, +Settings, +Type, +Entry, +Clause,
            +IntervalsIn, -IntervalsOut, -Count)
\end{progenv}
unifying \prog{IntervalsOut} and
and \prog{Count} with the number of examples within \prog{IntervalsIn}
that are covered by \prog{Clause}. The second branch of
\prog{induce/1}
uses \prog{prove_cache/8} in the same way as
\prog{induce/0} does in Aleph, localising the changes that need to be
made to introduce MPI in Aleph, and keeping them appart from the parts of
Aleph that implement the search itself.

The code of Aleph's \prog{prove_cache/8} is moved to
\prog{prove_cache_local/8}, and the new \prog{prove_cache/8} is
broadcasting the prove request
to the workers and collecting the results. This latter task consists
of accumulating the union of all the partial coverage list into a total
coverage list, and is performed quasi-asynchronously. In particular, it
is implemented as a loop of \prog{mpi_receive/3}'s without any sender
specified. As soon as a partial list is sent from any of the workers,
the master calculates its union with the cover-list accumulator and
reiterates to wait for the next partial result.

The workers are using \prog{prove_cache_local/8} (which contains the
code of the original \prog{prove_cache/8} predicate) to do the actual
proving and then transmit the list of successful intervals back to the
master.

\section{Testing Aleph/MPI}
\label{paral:results}
\label{paral:conclusions}

Given a programme that in some way processes or transforms its input
to produce output, any attempt to parallelise it would fall under one
of the two major brands of parallelism: \term{code-parallelism}, which
distributes the work that needs to be done to process each piece of
input, or \term{data-parallelism}, which distributes the input and
processes each individual piece of input sequentially. Since ILP
systems are programmes that transform an extensional definition of the
target predicate (that is, the examples) to an intensional definition,
the \quotes{input} of an ILP system is the examples and this paper
is describing a data-parallel version of Aleph.

It is immediately obvious that the choice between code-parallelism and
data-parallelism is dictated by the nature of the problem to be
solved: the interactions between the input data might be too dense to
allow for data-parallelism or it might not be possible to parallelise
the code of the process.
In the case of ILP, data-parallelism is only
speeding-up the evaluation of each hypothesised clause, which makes
this form of parallelism good for tasks where a significant amount of
the total computational cost is spend for clause evaluation, in other
words when there are large numbers of examples available.

\subsection{Learning the odd numbers}

In order to test Aleph/MPI, a very simple induction task was devised,
where Aleph had to construct a trivial theory from a large data set.
This was meant to simulate a situation where the computational cost
stems from the volume of the data-set rather than the size of the
search space.

The task chosen was that of learning the odd numbers. The definitions
of odd numbers and \quoted{small odd} numbers were included in the
background, so that the search algorithm would first try:
\begin{progenv}
target(N) :- small_odd(N).
\end{progenv}
and then discover the perfect-scoring theory:
\begin{progenv}
target(N) :- odd(N).
\end{progenv}
The data-set used consisted of the first 100 thousand natural
numbers\footnote{Obviously split down the middle between positives and
  negatives.}.

This setup was then used on the Beowulf Linux Cluster of the
\term{Centre for High Performance Computing and
  Visualisation}\footnote{See \url{http://www.rug.nl/hpc/} for more
  information} of the University of Groningen. This cluster consists
of 96 Pentium-4 workstations with 512~Mb of memory each and 16
double-processor Pentium-4 workstations, also with 512~Mb of memory.
The wall-times (as reported by the cluster's
job scheduler, and averaged over ten runs) versus the number of nodes
employed in the computation are given in the left column of
Table~\ref{fig:paral:results}.

Although the run-times shown in this table show an increase in
performance until Ahmdal's Law takes effect at 16 nodes, they do not
compare favourably with the performance of plain Aleph on one of the
nodes of the same cluster (ten runs averaging at 139.5 seconds, std.
dev.~1.3). This suggests that the time spent proving the examples
does not outweigh the message transmission costs involved in
parallelising the example-proving phase.
There are two ways in which the
example-proving phase might be more computationally expensive; the
background knowledge might be more complex so that proving individual
examples becomes more expensive, or there might be more examples.

\begin{table}

\begin{centre}
\begin{tabular}{|r|c|c|}
\hline
Nodes & Light (secs) & Heavy (secs) \\
\hline
8     & 991.9        & 9352.0 \\
10    & 419.4        & 8773.8 \\
12    & 270.3        & 8546.8 \\
16    & 168.0        & 8406.3 \\
24    & 145.2        & 8313.0 \\
32    & 147.0        & 8281.0 \\
48    &              & 8254.0 \\
64    &              & 8253.0 \\
\hline
\end{tabular}
\end{centre}

\caption{Wall-time performance versus number of nodes for
the two \texttt{target(N):-odd(N)} experiments.}
\label{fig:paral:results}
\end{table}

To test the first hypothesis, the definition of the \term{odd/1}
predicate above has been adjusted to simulate the computational cost of
a heavier, more difficult to prove background:
\begin{progenv}
odd(N) :- 
  once(sleep(0.01)),
  M is N mod 2, M =:= 1.
\end{progenv}
and the experiment was repeated, with the new run-times shown in
the right column of Table~\ref{fig:paral:results}. It is immediatelly
obvious that the
heavier background has made the benefits gained from each extra
processor smaller, and is still less efficient than vanilla Aleph
(7165~seconds), but on the other hand the effect of Amdahl's Law
manifests itself later, since with 64 nodes there is still some minor
gain whereas with the original experiment adding nodes stopped paying
off somewhere between 24 and 32 nodes.

The reason for this result is that a difficult background theory has
an impact on the saturation stage as well as the proving stage, since
the ground values encountered in the example that is being saturated
are tried on the background predicates, before these predicates
can be used as their minimal generalisation. Consequently, `heavier'
background predicates make the learning process lengthier, but does
not increase the fraction of the total time spent during the example
proving phase.

\subsection{Data-Parallelism Vs. Or-Parallelism}

Predicates consist of multiple clause which represent multiple ways
for the predicate to be satisfied; individual clauses might also
include explicit OR operators in their bodies; finally there might be
multiple ways to instantiate the variables in a clause's literals. All
these situations constitute \term{choicepoints} for a Prolog engine,
where one of many alternative paths through the proof search space has
to taken. The Prolog backtracking mechanism will exhaust each path
before backtracking to the last choicepoint and trying the next
option, and so on until the goal is satisfied or there are no choices
left and the goal fails.

An Or-parallel Prolog implementation is one that tries all the clauses
at a choicepoint in parallel. Or-parallelism can be used to divide the
search space of the ILP algorithm in `sectors' which will be searched
in parallel by the nodes of the machine, since the ILP algorithm's
search is actually the search for a proof: each time a literal is
added to the clause under construction it is picked from the `pool'
of literals provided by the bottom clause. In (very schematic) Prolog
this is implemented along the lines of this code fragment:
\begin{progenv}
add_one_literal(C, NextC):-
   bottom_clause( (Head:-Body) ),
   member(Lit, Body),
   append(Lit, C, NextC),
   is_good_clause(NextC).
\end{progenv}
which will backtrack back to \prog{member/2} each time
\prog{is_good_clause/1} is not satisfied, until all the member of the
bottom clause have been tried out. An Or-parallel Prolog
implementation would employ the nodes of a parallel machine to try out
all the ways to instantiate \prog{Lit} in \prog{member(Lit, Body)} in
parallel.

The data-parallelism described in this paper is, on the other hand, a
parallel implementation of the evaluation function employed by
\prog{is_good_clause/1} to decide whether to accept the
current clause or not.

It should, then, be noted that the computation expense discussed above
cannot be treated by
data-parallelism, since most of
the time is consumed in constructing candidate clauses and traversing the
search space, rather than the bottleneck being the large
amount of data against which each hypothesis needs to be tested.

\subsection{Conclusions}

The most important conclusion drawn from the above is the confirmation
of the fact that data-parallelism is mostly applicable to situations
where the bottleneck is the volume of the data rather than the
inherent computational complexity of the task.
In this case
it also offers the added advantage of spreading the examples among the
nodes, lowering the memory requirements per node for being able to fit
the examples in the machine.

There are, however, also situations where this form of parallelism is
not appropriate, since the bottleneck is the size of the search space
and the lack of reliable heuristics, rather than the effort of proving
the examples. In these cases it would be preferable to parallelise the
traversal of the search space, so that each node constructs and
evaluates a sub-set of the clauses hypothesised before a good clause
is identified.

\bibliography{ilp03-arxiv}

\begin{thebibliography}{8}
\providecommand{\natexlab}[1]{#1}
\providecommand{\url}[1]{\texttt{#1}}
\expandafter\ifx\csname urlstyle\endcsname\relax
  \providecommand{\doi}[1]{doi: #1}\else
  \providecommand{\doi}{doi: \begingroup \urlstyle{rm}\Url}\fi

\bibitem[Forum(1995)]{mpi:mpi}
MPI Forum.
\newblock \emph{{MPI}: A Message-Passing Interface Standard}.
\newblock \url{http://www.mpi-forum.org/docs/docs.html}, June 1995.

\bibitem[Forum(1997)]{mpi:mpi2}
MPI Forum.
\newblock \emph{{MPI-2}: Extensions to the Message Passing Interface}.
\newblock \url{http://www.mpi-forum.org/docs/docs.html}, July 1997.

\bibitem[Gropp and Lusk(1996)]{gropp:1996}
William~D. Gropp and Ewing Lusk.
\newblock \emph{User's Guide for {MPICH}, a Portable Implementation of {MPI}}.
\newblock Mathematics and Computer Science Division, Argonne National
  Laboratory, 1996.
\newblock URL \url{http://www-unix.mcs.anl.gov/mpi/mpich/docs.html}.
\newblock ANL-96/6.

\bibitem[Muggleton(1995)]{muggleton:1995}
Stephen Muggleton.
\newblock Inverse entailment and {Progol}.
\newblock \emph{New Generation Computing}, 13:\penalty0 245--286, 1995.
\newblock URL \url{ftp://ftp.cs.york.ac.uk/pub/ML_GROUP/Papers/InvEnt.ps.gz}.

\bibitem[Muggleton and De~Raedt(1994)]{muggleton:1994}
Stephen Muggleton and Luc De~Raedt.
\newblock {Inductive Logic Programming}: Theory and methods.
\newblock \emph{Journal of Logic Programming}, 19\penalty0 (20):\penalty0
  629--679, 1994.
\newblock URL \url{ftp://ftp.cs.york.ac.uk/pub/ML_GROUP/Papers/lpj.ps.gz}.
\newblock Updated version of technical report CW 178, May 1993, Department of
  Computing Science, K.U. Leuven.

\bibitem[Pacheco(1998)]{pacheco:1998}
Peter~S. Pacheco.
\newblock \emph{A User's Guide to {MPI}}, March 1998.
\newblock URL \url{ftp://math.usfca.edu/pub/MPI/}.

\bibitem[Srinivasan(2002)]{srinivasan:aleph}
Ashwin Srinivasan.
\newblock \emph{The {A}leph Manual}.
\newblock
  \url{http://www.comlab.ox.ac.uk/oucl/research/areas/machlearn/Aleph/}, Last
  update: Nov 21, 2002.

\bibitem[{Universidade do Porto}(Last update: April 2006)]{costa:yap}
{Universidade do Porto}.
\newblock \emph{{YAP} Prolog}.
\newblock \url{http://yap.sourceforge.net/}, Last update: April 2006.

\end{thebibliography}

\end{document}